\documentclass{article}
\usepackage{graphicx}
\usepackage{subcaption}
\usepackage{booktabs}
\usepackage{graphicx}
\usepackage[table]{xcolor}

\usepackage{cite}

\usepackage[left=3cm, right=3cm, top=4cm, bottom=4cm]{geometry}

\definecolor{rowred}{RGB}{245,200,200}
\definecolor{rowgreen}{RGB}{200,235,200}
\definecolor{roworange}{RGB}{245,215,185}

\title{Opportunities of Self Supervised Learning for GNSS: Evaluation of a Deep Learning-Enhanced PVT Algorithm}
\author{Thomas Barbero$^{1}$, 
Bertrand Ekambi$^{1}$
\\
\small
$^{1}$Abbia GNSS Technologies, Toulouse, France
\\
\thanks{Presented at the European Navigation Conference (ENC) 2026, Vienna, Austria. The final published version will be available via DOI upon publication.} 
\small
Presented at the European Navigation Conference 2026
}

\date{}

\begin{document}

\maketitle

\begin{abstract}
This work proposes a Deep Learning Enhanced PVT algorithm to mitigate multipath interference in dense urban areas. A supervised objective jointly predicts range corrections and uncertainty, while a JEPA-based self-supervised pretraining stage improves representation quality. The algorithm is evaluated over diverse driving scenarios, substantially improving PVT accuracy, particularly for unseen harsh urban conditions. These results highlight the potential of unlabelled GNSS data to improve generalization performance.
\end{abstract}


\section{Introduction}
In urban environments, GNSS signals reflect off buildings before reaching receivers, producing delayed replicas that cause multipath (MP) interference. MP is a leading source of positioning errors in dense urban areas \cite{groves_portfolio_nodate}. Because it depends on the receiver’s surroundings, mitigating it is extremely difficult \cite{groves_portfolio_nodate}. Methods such as ray-tracing tackle the problem from a physical standpoint, combining 3D city models with propagation approximations  \cite{groves_portfolio_nodate}. However, simplifications of the environment and physics inevitably cause prediction errors; therefore, complementary GNSS-observable analysis increases robustness.

MP leaves a distinctive signature in GNSS observables. Deep Learning (DL) can exploit this signature and its spatial correlations to model MP degradations \cite{zhang_prediction_2021}. DL has notably been applied to uncertainty estimation and Non-Line-Of-Sight (NLOS) identification \cite{zhang_prediction_2021, sun_stacking_2022}. Yet, the deployment of DL-based methods remains challenging due to substantial variations of the GNSS observables with receiver hardware/software and the diversity of urban environments \cite{groves_portfolio_nodate, sun_stacking_2022}.
Increasing the volume and diversity of the training dataset improves generalization performance \cite{alzubaidi_survey_2023, barbero_toward_2026}. However, producing ground truth labels such as reference positions requires a dedicated testbed with a high-end reference receiver, making large-scale collection impractical.

Incorporating data from standalone receivers is therefore attractive but requires training methods that do not depend on ground truth information. For this purpose, we adopt Self-Supervised learning (SSL): Deep Neural Networks (DNN) are pretrained over large unlabelled datasets and finetuned over task specific labelled data \cite{alzubaidi_survey_2023}. Adaptation of SSL to GNSS observables is currently under-explored, especially the extensive pretraining of DNN for general-purpose feature extraction. Additionally, the highly effective Masked Modelling pretraining objective has not been adapted to GNSS. 

We address this gap by adapting the Joint Embedding Predictive Architecture (JEPA) \cite{assran_self-supervised_2023} to GNSS observables. The pretrained Encoder serves as feature-extracting backbone for a MP prediction architecture trained with a heteroscedastic regression objective \cite{kendall_what_2017} that jointly predicts code corrections and uncertainty. The same supervised objective is also used to train models from scratch, enabling a direct comparison between SSL pretraining and Supervised training only. Predicted corrections and uncertainty are integrated into a Weighted Least Squares (WLS) estimator. The overall method is evaluated through the accuracy of the resulting enhanced PVT solution. Models are trained and evaluated using the combination of two open-source datasets \cite{urbannav, suzuki_open-source_2025} and 24h of driving data; covering diverse urban environments. We report substantial accuracy gains for our DL-Enhanced PVT algorithm (DLE-PVT). Moreover, our analysis show that SSL encourages DNN to rely on robust correlations to make its predictions, yielding predictions that remain reliable under harsh and unseen conditions.\\

Our contributions are threefold: (1) We adapt the JEPA self-supervised framework to GNSS observables, enabling future large-scale pretraining on unlabelled data from standalone receivers. (2) We employ a heteroscedastic supervised objective that jointly predicts code corrections and residual error variance; providing MP estimate and uncertainty measure. (3) We integrate DNN predictions into a WLS PVT estimator and demonstrate accuracy gains in diverse urban environments.

\section{Deep Learning Enhanced PVT Algorithm}

The proposed DLE-PVT algorithm is decomposed into two main components: a multipath characterization module and a Multipath-Aware Positioning Algorithm (MAPA); both illustrated in Figure 1 (b). Training of the DNNs is entirely separated from DLE-PVT. This section details the DLE-PVT algorithm; while Section 3 details the training approaches.

\begin{figure*}[t]
    \centering
    \begin{subfigure}[t]{\textwidth}
        \centering
        \includegraphics[width=0.8\linewidth]{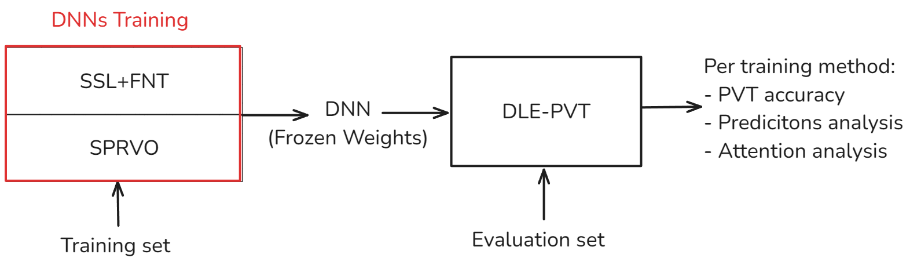}
        \caption{Trained models are embedded into DLE-PVT to produce evaluation results per training method.}
        \label{fig:method_a}
    \end{subfigure}

    \vspace{6pt}

    \begin{subfigure}[t]{\textwidth}
        \centering
        \includegraphics[width=0.8\linewidth]{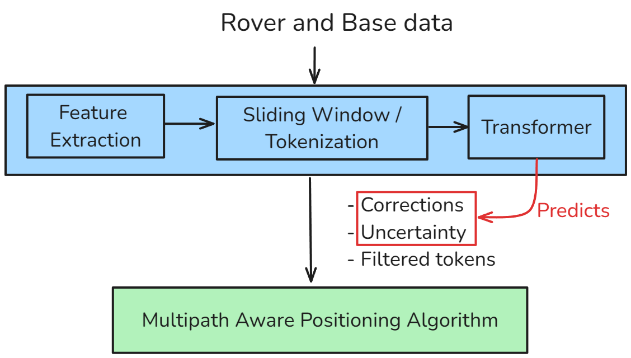}
        \caption{During evaluation, the predicted corrections and uncertainties are integrated into the multipath-aware positioning algorithm.}
        \label{fig:method_b}
    \end{subfigure}

    \caption{Methodology overview.}
    \label{fig:methodology}
\end{figure*}

\subsection{Features Extraction}
The features extraction step produces two categories of features: Contextual Features and Multipath Indicators. An initial PVT is computed using a WLS similar to the MAPA without corrections and with a weighting based on CN0 and elevation following \cite{herrera_gogps_2016}. Both the initial WLS and MAPA use Single Differenced (SD) code observable and Doppler Shift to estimate PVT. The SD operation using a nearby base station largely removes atmospheric and satellite ephemeris/clock error sources from the rover measurements \cite{zhang_improving_2023}. Only the main frequency of each constellation is used. The estimated velocity is transformed to East North Up (ENU) coordinates and used as Contextual Features. The Contextual features collectively serve to provide spatial (azimuth/elevation/velocity) and temporal awareness; plus the identification of individual satellites (constellation and ID).

MP Indicators are features with established correlation to MP such as Carrier-to-Noise Ratio (CN0) \cite{groves_portfolio_nodate}. Moreover, we employ a range-rate consistency metric following \cite{sun_stacking_2022}, which grows with the discrepancy between the difference of successive pseudorange measurements and Doppler-derived range-rates. Finally, the Doppler and SD Code residuals from the initial WLS are also used as MP Indicators. 

The extracted features for a same satellite at a given timestep are gathered into a 12d vector, called token. A sequence of tokens is formed with a sliding-window approach: new tokens are produced at measurements reception, and the window retains the 100 most recent tokens. Each feature is standardized using the mean and standard deviation computed over the training dataset. Tokens are excluded if any feature is missing or falls outside the empirically defined inclusion ranges listed in Table 1. 

\begin{table}[ht]
\centering
\caption{Summary of the extracted features and their respective inclusion range.}
\label{tab:features}
\begin{tabular}{|l|c||l|c|}
\hline
\multicolumn{2}{|c||}{\textbf{Contextual Features}} &
\multicolumn{2}{c|}{\textbf{MP Indicators}} \\
\hline
\textbf{Name} & \textbf{Inclusion Range} &
\textbf{Name} & \textbf{Inclusion Range} \\
\hline
Azimuth & ]0,360] & C/N$_0$ Rover & $>5$ \\
Elevation & ]0,90] & Rover/Base C/N$_0$ Ratio & N/A \\
Satellite ID \& Constellation & N/A &
Range-Rate Consistency & ]$-10$,10[ $\backslash$ \{0\} \\
GPS Time & N/A &
Doppler Residuals & ]$-10$,10[ $\backslash$ \{0\} \\
Velocity (ENU) & N/A &
SD Code Residuals & ]$-150$,150[ $\backslash$ \{0\} \\
\hline
\end{tabular}
\end{table}

\subsection{Deep Learning Modelling}
Sequences are passed as input to a Transformer \cite{vaswani_attention_2023} which predicts code correction and uncertainty for each token. The Transformer architecture is composed of four successive components: (1) Features/Token projection; (2) Encoder with Multi-Head Self Attention (MHSA); (3) Decoder with MHSA and Cross Attention (CA) over the Encoder’s last representations; (4) Multi-Layer-Perceptron (MLP)-based Prediction Head.

The Encoder serves as a general feature extracting backbone, while the decoder extracts specialized correlations during the finetuning or pretraining stage. CA re-queries the robust representations produced by the encoder. Predictions are produced jointly for all tokens in the window rather than autoregressively. The Transformer follows the original formulation \cite{vaswani_attention_2023} and is implemented using Pytorch. The architecture consists of 4 encoder and 2 decoder blocks with 4 attention heads each, an embedding dimension of 32 and a feed-forward dimension of 128. Encoder dropout is 0.1; decoder dropout is 0 during SSL and 0.1 during finetuning. The prediction head is a 2-layer MLP with a hidden dimension of 64 and ReLU activation. The total parameter count is approximately 100k.

 Transformers require positional information for each token, as the attention mechanism is invariant to token order. We derive this positional embedding from the Contextual Features. Time and satellite ID are first mapped to relative indices. Constellation and satellite ID are then passed through embedding layers, while time and velocity are processed by linear projections. Azimuth and Elevation are first combined into a trigonometric encoding [sin elev, cos elev, sin azim, cos azim] before linear projection, ensuring continuity across the azimuth wrap-around at 360°. MP Indicator features are jointly passed through a single linear projection. All projections produce vectors of the Transformer's embedding dimension, which are summed to form the token embedding fed to the encoder.
 
\subsection{Multipath-Aware Positioning Algorithm}
The estimated state at time $t$ is
\[
\mathbf{x}_t =
\left\{
\mathbf{x}_t,\,
\mathbf{v}_t,\,
\Delta\Delta T_t,\,
(\Delta T)_{t,\mathrm{sys}}
\right\},
\]
corresponding to the rover position, velocity, clock drift, and the GPS time bias between the rover and the base station for each constellation. The system is formulated as a factor graph and solved with GTSAM using a Levenberg--Marquardt optimizer \cite{dellaert_factor_2012}. The factor graph is constructed over a single epoch, making the estimation equivalent to a WLS solver. We adopt this formulation to enable extension to multi-epoch estimation in future work.

DNN code corrections are inverse-standardized using the mean and standard deviation of the reference corrections from the training set (see Section 3) before being applied to the SD code observables. The DNN predicts log-uncertainty for numerical stability; the prediction is therefore exponentiated and rescaled by the same standard deviation (no mean offset, as uncertainty is a scale parameter) to recover a standard deviation in meters, which is used as the noise model for each associated SD code factor. Tokens excluded from DNN processing are not removed from the MAPA; instead, their associated SD code observables are retained with a fixed standard deviation of 500m, which diminishes their contribution while preserving satellite geometry.

\section{Deep Neural Networks Training}
This work evaluates two distinct training approaches: Supervised Only Training (SPVO) and SSL followed by Supervised finetuning (SSL+FNT). Both training routes produce models with identical final architectures that predict code corrections and uncertainties. The supervised objective is shared between the two routes: it is the unique training method for SPVO and is employed during finetuning for SSL+FNT. The training procedure is entirely separated from the DLE-PVT algorithm. This section details both training objectives, before explaining the two distinct training routes. 

\subsection{Self-Supervised Training}
To capture the underlying spatial structure of MP, Masked Modelling is employed as the SSL training objective: a part of the input sequence is masked and the model is trained to predict the masked tokens from the non-masked ones. We design a spatially-motivated masking strategy: for each sequence, a continuous azimuthal sector covering 25\% of the sky is randomly selected. Low elevation tokens ($<$ 40°) within this sector are masked. Reconstruction of tokens can be done in the input space (as in Auto-Encoders \cite{alzubaidi_survey_2023, barbero_toward_2026}) or representation space. The latter is more adapted to GNSS features which are noisy, and which exact reconstruction could confuse the model. For this reason, we employ the JEPA framework \cite{assran_self-supervised_2023}. Input sequence is fed to two distinct branches, which generate embeddings for each token: (1) Target Branch produces target embeddings from the full sequence without masked tokens; (2) Prediction Branch produces predicted embeddings from the full sequence with masked tokens. For masked tokens, MP Indicators projection is replaced by a learnt MASK embedding; but positional embedding is retained.

The loss is measured as the discrepancy between target and predicted embeddings for the masked tokens only. Two-branch architectures are prone to representation collapse without regularization. Instead of the Exponential Moving Average regularization, we add VICReg \cite{bardes_vicreg_2022} regularization terms to directly constrain the distribution of the embeddings. The Encoder is identical for the two branches. VICReg combines three losses components on batches of predicted and target embeddings: (1) Invariance: Mean Squared Error between target and predicted embeddings (for the masked tokens); (2) Variance: standard deviation of each embedding dimension across the batch is pushed toward 1; (3) Covariance: off-diagonal covariances between embedding dimensions across the batch are pushed toward 0. Loss terms are empirically tuned at 1, 1, and 0.01 respectively. Variance and Covariance regularization are applied to the Encoder representations directly. The covariance term excludes token pairs from the same satellite within a given sequence, as embeddings are expected to be tightly correlated in this case.

\subsection{Supervised Training}
The Supervised training objective uses a reference MP approximation to jointly guide uncertainty and correction predictions. This ground truth information is derived from SD Code observables by subtracting the geometric component (computed from the reference trajectory) and the clock component (which is estimated by a robust factor graph over the entire trajectory). The residual error is expected to approximate MP-induced delay since reference position is highly accurate and clock states are constrained by temporal coherency factors and robust M-Estimators [9]. We employ a heteroscedastic regression loss [7] that maximizes the likelihood of the residual error under a Laplace distribution. The Laplace distribution is preferred over a Gaussian one to reduce sensitivity to outliers. Equation 1 describes our loss function, where $\widehat{\log \sigma_{\mathrm{std}}}$ and $\widehat{\mathrm{MP}}$ respectively denote the log-uncertainty and code correction predictions, and $\mathrm{MP}$ denotes reference MP.

\[
\mathcal{L}
=
\frac{\left|\mathrm{MP}-\widehat{\mathrm{MP}}\right|}
{\exp\!\left(\widehat{\log \sigma_{\mathrm{std}}}\right)}
+
\widehat{\log \sigma_{\mathrm{std}}}.
\]

\subsection{Training Methodology}
SPRVO training is straightforward: the model is trained from scratch using the supervised objective only. For SSL+FNT, the model is first pretrained using the SSL objective. The encoder weights are then transferred to the finetuning stage, whereas the decoder and prediction head weights are reinitialized. Encoder weights are frozen, while the rest of the architecture is trained from scratch on the supervised objective. The decoder architecture is unchanged between SSL and finetuning; only its weights are reinitialized. The prediction head’s output dimension differs between stages: 2 during supervised training (correction + uncertainty) and matching the embedding dimension during SSL. Finally, during training, sequences are picked randomly instead of using a sliding window approach.

Models use the Xavier uniform Initialization for all weights except for embedding layers which are initialized from a normal distribution \cite{glorot_understanding_2010}. Weight decay is applied to all parameters except embeddings and layer norm parameters. Optimisation uses AdamW \cite{kingma_adam_2017}, the learning rate follows a linear warm-up and a cosine decay to 10\% of its base value. 

\begin{table}[ht]
  \centering
  \caption{Summary of the different training phases hyper-parameters.}
  \label{tab:hyperparams}
  \begin{tabular}{lccccc}
    \toprule
    & \textbf{Base Learning Rate} & \textbf{Weight Decay} & \textbf{Warmup} & \textbf{Epochs} & \textbf{Batch Size} \\
    \midrule
    Pretraining & 1e-4 & 0.04 & 30\% & 50  & 1000 \\
    Finetuning  & 3e-4 & 0.01 & 10\% & 30  & 500  \\
    SPRVO       & 3e-4 & 0.01 & 70\% & 80  & 500  \\
    \bottomrule
  \end{tabular}
\end{table}

\section{Empirical Setting}
The following section details the empirical setting by explaining the employed datasets, the evaluation strategies and the trained/evaluated DNN.

\subsection{Datasets}
Diverse urban driving scenarios were gathered from a proprietary data capture and the PPC and UrbanNav datasets \cite{urbannav, suzuki_open-source_2025}. All datasets use mid-range u-blox receivers covering Galileo, GPS, BeiDou and Glonass constellations. Our proprietary data capture consists of 24 hours of driving scenarios in Toulouse, France. Light Urban and Highway scenarios are grouped as Circuit \#1 (8 scenarios, ~1.35M tokens), while Deep Urban scenarios with intense shadowing from roadside trees are grouped as Circuit \#2 (4 scenarios, ~600k tokens). We used a testbed composed of a u-blox F9R receiver for test and a Septentrio AsterX Sbi3Pro+ GNSS-INS receiver for reference. Both receivers share a Septentrio antenna. The reference trajectory was refined through post-processing with the Qinertia tool.

PPC and UrbanNav are established open-source datasets covering diverse harsh urban conditions in Tokyo and Hong Kong. Two PPC scenarios were discarded due to extended GNSS outages. The remaining ones are grouped as OS (9 scenarios, ~130k tokens). 

\subsection{Evaluation Protocols}
A strict scenario-level split is applied to prevent data leakage: any scenario used for training or validation of the supervised task is excluded from the test set. Circuit \#1 scenarios form the basis of the training set optionally diversified with subsets of the other two datasets depending on the training configuration (see Section 4.3). Held-out Circuit \#1 scenarios are designated In Distribution (ID). Circuit \#2 scenarios, which differ slightly from the main training distribution are designated Slight OOD. The open-source datasets, which differ substantially from the dominant training distribution are called Heavy OOD.

Trained models are evaluated through their integration into the DLE-PVT algorithm, which results are reported per training method and scenario category. We compare results to a baseline based on the initial WLS PVT, with the same down-weighting of filtered tokens as in the MAPA. Accuracy is measured as the average of the 50th and 95th percentiles of 3D positioning error; following \cite{suzuki_open-source_2025}. This metric captures both typical and degraded performances in a single score. To improve results consistency, each training method is repeated under identical conditions with different random seeds; per-epoch positioning error are pooled across run before computing the 50th and 95th percentiles.

\subsection{Experiments and Training configurations}
We propose two experiments with distinct training set compositions. The first compares SPRVO and SSL+FNT on a fixed Circuit \#1 training set of ~1M tokens, isolating the impact of SSL pretraining. The second focuses on SSL+FNT and evaluates the impact of injecting portions of the other datasets to the pretraining (+~600k tokens) or finetuning stage (+~150k tokens), or to both. Scenarios used for finetuning are removed from evaluation. The ones used only for pretraining remain available for test, as SSL objective does not directly target the downstream task. For this second setting, the model is scaled up to ~500k parameters to accommodate the increased data diversity. We use 5 training runs for the first experiment and 3 for each configuration of the second; for SSL+FNT, only pretraining is repeated and finetuning is done once for each pretrained model.

\section{Empirical Findings}
We detail results of two experiments: a first one isolates the impact of pretraining; and a second one examines the impact of the diversification of the training set.

\begin{table}[ht]
  \centering
  \caption{Average of the 50\textsuperscript{th} and 95\textsuperscript{th} percentiles of 3D Positioning Error per method and dataset.}
  \label{tab:3d_positioning_error}
  \begin{tabular}{lccc}
    \toprule
    & \textbf{ID} & \textbf{Slight OOD} & \textbf{Heavy OOD} \\
    \midrule
    Baseline  & 4.98         & 9.16         & 28.76        \\
    SPRVO     & \cellcolor{rowgreen} 2.85         & \cellcolor{rowgreen} \textbf{7.28} & \cellcolor{rowgreen} 21.26        \\
    SSL+ FNT  & \cellcolor{rowgreen} \textbf{2.64} & \cellcolor{rowgreen} 7.72         & \cellcolor{rowgreen} \textbf{18.96} \\
    \bottomrule
  \end{tabular}
\end{table}

\subsection{Impact of Pretraining}

\subsubsection{Performance of the DLE-PVT Algorithm}
Table 5 shows that DNN predictions substantially and consistently improve the accuracy of the PVT algorithm over the baseline; regardless of the training method. For ID and Slight OOD scenarios, the baseline is already accurate, and both training methods yield similar positioning performances. SPRVO slightly outperforms SSL+FNT for the Slight OOD dataset, which we attribute to shadowing features underrepresented in pretraining data. The simpler SPRVO features happen to be more effective to these conditions (see Section 5.1.2). For Heavy OOD, the harsh urban conditions cause the baseline to degrade severely. For those conditions, SSL+ FNT outperforms SPRVO substantially. SSL+FNT increases resilience to degraded initial PVT conditions. The next subsection analyses the differences in predictions for the two training methods.

\subsubsection{Impact of Pretraining at Predictions Level}
We aggregate results for the different training runs on the Heavy OOD scenarios, where the gap in performance for the training methods is the largest. Figure 2 (a) shows that the distribution of weighted range residual error for the SPRVO method exhibits heavy tails, indicating that a substantial part of MP is not mitigated at the range level. The tails almost disappear for SSL+FNT which yields more coherent predictions under harsh conditions. 

\begin{figure*}[!htbp]
    \centering
    \begin{subfigure}[t]{\textwidth}
        \centering
        \includegraphics[width=0.6\linewidth]{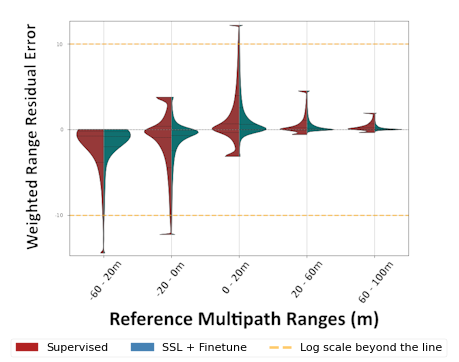}
        \caption{Distribution of Weighted Range Residual Error with Respect to bins of Reference Multipath.}
        \label{fig:distrib}
    \end{subfigure}

    \vspace{6pt}

    \begin{subfigure}[t]{\textwidth}
        \centering
        \includegraphics[width=0.8\linewidth]{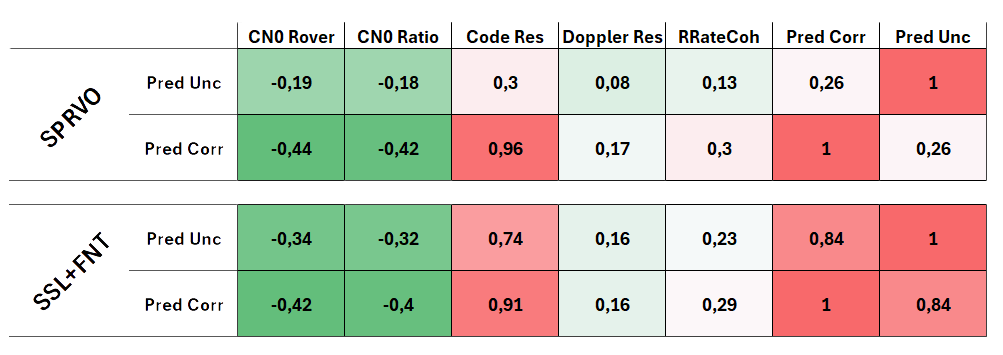}
        \caption{Correlation of Predictions and Features.}
        \label{fig:correlation}
    \end{subfigure}

    \caption{Range/Prediction Level analysis.}
    \label{fig:range_analysis}
\end{figure*}

Figure 2 (b) shows differences between the correlations of predictions and MP Indicators. First, the correction prediction correlates strongly with the initial WLS SD code residual in both cases, but more for SPRVO (0.96 vs 0.91). Secondly, the uncertainty prediction correlates meaningfully with several features for SSL+FNT, while it remains loosely correlated for SPRVO. We propose the following explanation: SPRVO converges to a near-identity mapping from code residuals to corrections, which holds well when the initial PVT is accurate. This shortcut encourages the model to minimize uncertainty predictions to further reduce the loss, but uncertainty becomes poorly grounded. As the initial PVT degrades for Harsh OOD conditions, the learnt features are no longer reliable. Conversely, the SSL encoder has no incentive to preserve the exact residual values, but learns representations grounded in spatial correlations. The decoder must therefore predict corrections from these richer representations, which requires to use grounded uncertainty predictions to mitigate MP effectively. As a result, SSL+FNT predictions remain coherent under harsh conditions. The next subsection shows the spatial understanding of our SSL+FNT models.

\subsubsection{Attention Analysis}
We investigate the attention mechanism which is at the core of the Transformer architecture \cite{vaswani_attention_2023}. Attention enables the exchange of information between tokens, weighted by pairwise attention scores. The spatial distribution of high attention scores reveals where the models find relevant information. Figure 3 shows the mean attention score per layer, averaged across attention heads, in a relative sky plot: elevation is absolute, while azimuth is relative between attending and attended tokens (0°=same azimuthal sector, 180°=opposite one). High attention scores (=key regions) appear in yellow colour. Results are shown for a randomly selected SSL+FNT model on UrbanNav scenarios. 

Attention is spatially structured. In the encoder, the first two layers mostly attend to high elevation. The next two layers attend to low elevation tokens, where MP is more intense. The last layer of the decoder, which is the closest to the predictions, attends intensely to the current token’s own azimuthal sector; encouraging spatially coherent predictions for same sky regions.

\newsavebox{\imgA}
\newsavebox{\imgB}
\newlength{\gapwidth}
\newlength{\ratioA}
\newlength{\ratioB}
\newlength{\imgAwidth}
\newlength{\imgBwidth}
    
\begin{figure*}[!htbp]
    \centering
    \savebox{\imgA}{\includegraphics[width=1cm]{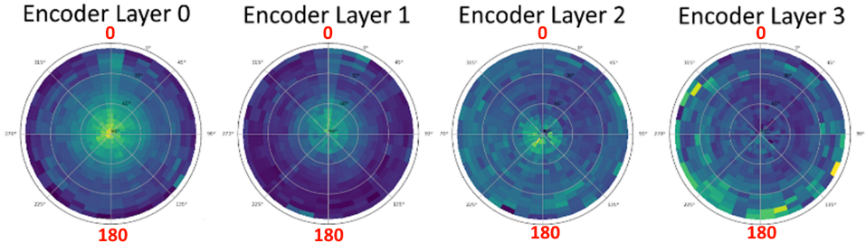}}
    \savebox{\imgB}{\includegraphics[width=1cm]{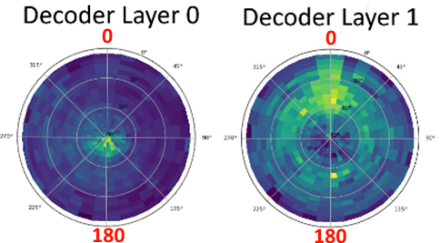}}
    \setlength{\gapwidth}{0.04\textwidth}
    \setlength{\ratioA}{\dimexpr\wd\imgA*1000/\ht\imgA\relax}
    \setlength{\ratioB}{\dimexpr\wd\imgB*1000/\ht\imgB\relax}
    \setlength{\imgAwidth}{\dimexpr(\textwidth-\gapwidth)*\ratioA/(\ratioA+\ratioB)\relax}
    \setlength{\imgBwidth}{\dimexpr\textwidth-\gapwidth-\imgAwidth\relax}
    \begin{subfigure}[t]{\imgAwidth}
        \centering
        \includegraphics[width=\linewidth]{figures/image3_encoder.png}
        \caption{Encoder.}
        \label{fig:encoder}
    \end{subfigure}
    \hfill
    \begin{subfigure}[t]{\imgBwidth}
        \centering
        \includegraphics[width=\linewidth]{figures/image3_decoder.png}
        \caption{Decoder.}
        \label{fig:decoder}
    \end{subfigure}
    \caption{Average attention score per relative sky plot bin for the layers of (a) Encoder and (b) Decoder.}
    \label{fig:attention}
\end{figure*}

\subsection{Increasing Training dataset diversity}
This second experiment evaluates the impact of diversifying training data on performance, reported in Table 6. Diversity is added during pretraining, finetuning or both. Because the test scenarios differ from those of Section 5.1, Tables 5 and 6 are not comparable.

\begin{table}[ht]
  \centering
  \caption{Average of the 50\textsuperscript{th} and 95\textsuperscript{th} percentiles of 3D Positioning Error per pretraining and finetuning data combination.}
  \label{tab:data_combinations}
  \begin{tabular}{lccc}
    \toprule
    \textbf{Pretraining Data \textbar{} Finetuning Data} & \textbf{ID} & \textbf{Slight OOD} & \textbf{Heavy OOD} \\
    \midrule
    Basis \textbar{} Basis
      & 2.55
      & 9.00
      & 20.06 \\
    Basis \textbar{} \textbf{Mixed}
      & \cellcolor{rowred} 2.57
      & \cellcolor{rowred} 9.16
      & \cellcolor{rowgreen} \textbf{18.67} \\
    \textbf{Mixed} \textbar{} Basis
      & \cellcolor{rowgreen} \textbf{2.55}
      & \cellcolor{rowgreen} \textbf{8.59}
      & \cellcolor{rowgreen} 19.70 \\
     Mixed \textbar{} Mixed
      & \cellcolor{rowred} 2.68
      & \cellcolor{rowgreen} 8.82
      & \cellcolor{rowgreen} 18.71 \\
    \bottomrule
  \end{tabular}
\end{table}

Two contrasting trends emerge. Slight OOD performance improves most when diversification is applied to the pretraining stage. Heavy OOD shows the opposite pattern, diversification at the fine-tuning stage drives the largest improvement, while pretraining diversification provides negligible benefits in comparison. Diversifying pretraining allows the encoder to capture representations covering a broader range of conditions, which enable Slight OOD gains: richer representations for shadowed conditions transfer effectively to improved predictions. H

owever, diverse representations alone are insufficient when the labelled set is too narrow, as the decoder cannot learn to map this diversity to nuanced predictions. Heavy OOD illustrates this case: the encoder’s representations of harsh conditions are adequate, but the decoder has not been exposed to labelled examples of extreme MP magnitudes. Both labelled and unlabelled data diversity are currently limiting. We hypothesize that after modest additional labelled data collection, scaling the unlabelled pretraining set alone should drive further generalization gains.

\section{Conclusions}
This work proposes a DLE-PVT algorithm that substantially improves positioning accuracy, notably under unseen conditions. The supervised and self-supervised training methods were found to be effective, especially when employed together: the supervised objective predicts code corrections and uncertainty, while the SSL pretraining stage improves the coherence of predictions through higher-quality representations and reliance on robust features. Future work will explore whether scaling unlabelled data alone can drive further generalization gains. More fundamentally, multipath is governed by physics and geometry. Self-Supervised Learning anchors deep neural networks to this physical reality.

\bibliographystyle{IEEEtran}
\bibliography{ENC2026}

\end{document}